\documentclass{article}
\usepackage{spconf,amsmath,graphicx } 
\usepackage{xcolor}
\usepackage{multirow}
\usepackage{multicol}
\usepackage{flushend}
\usepackage[ruled,vlined]{algorithm2e}
\usepackage[skip=0pt]{caption} 
\BeforeBeginEnvironment{figure}{\vskip-2ex}
\AfterEndEnvironment{figure}{\vskip-1ex}
\usepackage{enumitem}
\setlist{nosep, leftmargin=14pt}

\usepackage{mwe} 

\usepackage{fancyhdr}
\fancypagestyle{firstpage}{
  \fancyhf{}
  \setlength{\headheight}{20mm}
  \renewcommand{\topmargin}{-15mm}
  \fancyhead[C]{\vspace{5mm} \small \textit{Proc. IEEE ISBI 2023}, Cartagena de Indias, Colombia, Apr. 2023\\ © 2023 IEEE. Personal use of this material is permitted. Permission from IEEE must be obtained for all other uses, in any current or future media, including reprinting/republishing this material for advertising or promotional purposes, creating new collective works, for resale or redistribution to servers or lists, or reuse of any copyrighted component of this work in other works.}}

\newcommand{\etal}{\textit{et al.}}

\title{Quality-Adaptive  
Split-Federated Learning for Segmenting Medical Images with Inaccurate Annotations}
\name{Zahra Hafezi Kafshgari, Chamani Shiranthika, Parvaneh Saeedi, and Ivan V. Baji\'{c}
}
\address{School of Engineering Science, Simon Fraser University, Burnaby, BC, Canada}

\begin{document}

%
\maketitle
\begin{abstract}
SplitFed Learning, a combination of Federated and Split Learning (FL and SL), is one of the most recent developments in the decentralized machine learning domain. In SplitFed learning, a model is trained by clients and a server collaboratively.
For image segmentation, labels are created at each client independently and, therefore, are subject to clients' bias, inaccuracies, and inconsistencies. In this paper, we propose a data quality-based adaptive averaging strategy for SplitFed learning, called QA-SplitFed, to cope with the variation of annotated ground truth (GT) quality over multiple clients.
The proposed method is compared against five state-of-the-art model averaging methods on the task of learning human embryo image segmentation. Our experiments show that all five baseline methods fail to maintain accuracy as the number of corrupted clients increases. QA-SplitFed, however, copes effectively with corruption as long as there is at least one uncorrupted client.

\end{abstract}
\begin{keywords}
SplitFed learning, privacy, poor annotation, ground truth corruption, data heterogeneity 
\end{keywords}

\thispagestyle{firstpage}

%
\section{Introduction}
\label{sec:intro}
Decentralized learning decouples data from the computational infrastructure, and removes the need for sharing data, in both training and inference. Instead, features derived from the data at various levels of processing are shared among clients and servers. Besides benefits to privacy, decentralized learning could scale to the available resources, both human and computational, available at different locations. This makes it attractive for medical applications, where clients may be hospitals, clinics, and medical centers. SplitFed Learning (SFL) is a combination of two decentralized learning methods: Split Learning~\cite{vepakomma2018split}, and Federated Learning~\cite{mcmahan2017communication}. In FL, clients keep their own data, but must  train their local full models and send their model weights to the server for averaging~\cite{li2020review}. Hence, in FL, each client must have sufficiently powerful hardware to train its own model. SL methods shift the computational burden from the client to the server. In SL, the model is split into two or more parts, and training is done collaboratively by clients and a server communicating the activations and gradient updates of the split layers~\cite{SL}. SFL takes advantage of both FL and SL, and addresses their shortcomings~\cite{splitfed}. Specifically, in SFL, each client's data stays on the client and most of the computational burden is offloaded to the server. 

In this paper, we study SFL in the case where data annotation quality differs across clients. Specifically, we focus on embryo image segmentation, where ground truth segmentation accuracy may vary from client to client, depending on the expertise and human resources available at each client. We propose a model averaging procedure that is significantly more robust in maintaining the global accuracy to such variation in annotation quality compared to existing methods. Related work is reviewed in Section~\ref{sec:related}. Proposed method is presented in Section~\ref{sec:proposed}, followed by experiments in Section~\ref{sec:experiments} and conclusions in Section~\ref{sec:conclusions}.

\section{Related works}
\label{sec:related}

Different types of data heterogeneity could occur in the context of collaborative learning, e.g. unbalanced data~\cite{mcmahan2017communication}, different feature map distributions~\cite{splitAVG}, or different levels of annotation accuracies over clients~\cite{fedmix}. In the literature, these are commonly referred to as non-IID (Independent and Identically Distributed) scenarios.  
McMahan \etal~\cite{mcmahan2017communication} first introduced the concept of FL by averaging all local model weights to update a global model. Their model averaging procedure, called FedAVG, assigned a weight proportional to the number of data samples at each client.  
Several extensions to FedAVG exist in the literature. FedAVG-M is an optimized version of FedAVG that applies a momentum on the server optimizer to deal with Non-IID clients~\cite{FedAVGM}.
Shi \etal~\cite{shi2021fed} introduced Fed-ensemble to average \textit{k} randomly trained models in the federation step to reduce the variance of the global model.
Fang \etal~\cite{robust_FL} studied heterogeneous and noisy labels in FL. Noisy clients were created by symmetric and pair flip noise, and a public dataset was used to adjust the global model.  
For medical applications, accessing patients’ public data is difficult due to privacy concerns. Yang \etal~\cite{FL_noisyLabels} proposed a method in which noisy labels were created in a similar way to those in~\cite{robust_FL}. The average of the local feature vectors per client, called centriods, were shared with the server for model aggregation and label correction.
In~\cite{auto_robust_FL}, noisy labels were created by shuffling labels randomly and adding Gaussian noise to the input data. 
The weighted sum of empirical risks was then minimized for clients with regularization terms. 
Considering statistical data heterogeneity in FL, FedProx was proposed in~\cite{li2020federated} to stabilize the final model by a proximal term added to the optimization function. 

The idea of combining FL and SL into SplitFed (or SFL) was first introduced  by Chandra \etal~\cite{splitfed}.  
They proposed two versions of SplitFed: vanilla SplitFed (SplitFedv1) and U-shaped SplitFed (SplitFedv2). In vanilla SplitFed, labels are shared with the server, while in the U-shaped version, all data is kept private on the clients. Simple arithmetic averaging of model weights was used in both cases.  
In the healthcare domain, Poirot \etal~\cite{SFL_healthcare} first utilized U-shaped SplitFed in experiments on fundus photos and chest X-rays. Gawali \etal~\cite{comparison_FL_SL_SFL} reviewed FL, SL, and SplitFed approaches in the medical domain, and introduced SpltFedv3, which was a label sharing approach (similar to SpliFedv1). SpltFedv3 did not change the client-side model weights, and only applied the federation on the server-side model weights, while still using simple averaging. 

SplitAVG, a very recent work, was proposed to handle heterogeneous clients whose feature maps differ~\cite{splitAVG}. Feature maps from different clients are concatenated  
and fed to the server simultaneously. No averaging is performed on client-side models. 
Chen \etal~\cite{SFLenergy} studied another type of heterogeneity in terms of energy harvesting capability per client. Their proposed SplitFed algorithm was tested on CIFAR-10 and CIFAR-100 datasets. A weighted averaging strategy was proposed in~\cite{fedmix} that took advantage of labeled data for creating pseudo labels for unlabeled data. It used clients with better training quality when updating the global model. This method was applied on medical image segmentation task. 

In this paper,  a novel method for SplitFed model averaging, called QA-SplitFed, is proposed to address scenarios where not all clients' labels are annotated correctly. Model weights are updated twice in each global epoch based on the local and global behavior of all clients. QA-SplitFed is applied to training an embryo image segmentation model.  
To assess its robustness,  
a comparison with five baseline averaging methods is presented.

\vspace{-10pt}

\section{Proposed Method}
\label{sec:proposed}
\vspace{-5pt}
\subsection{Human embryo components segmentation model}
\begin{figure}[!tb]
    \centering
    \includegraphics[ width=8.8cm]{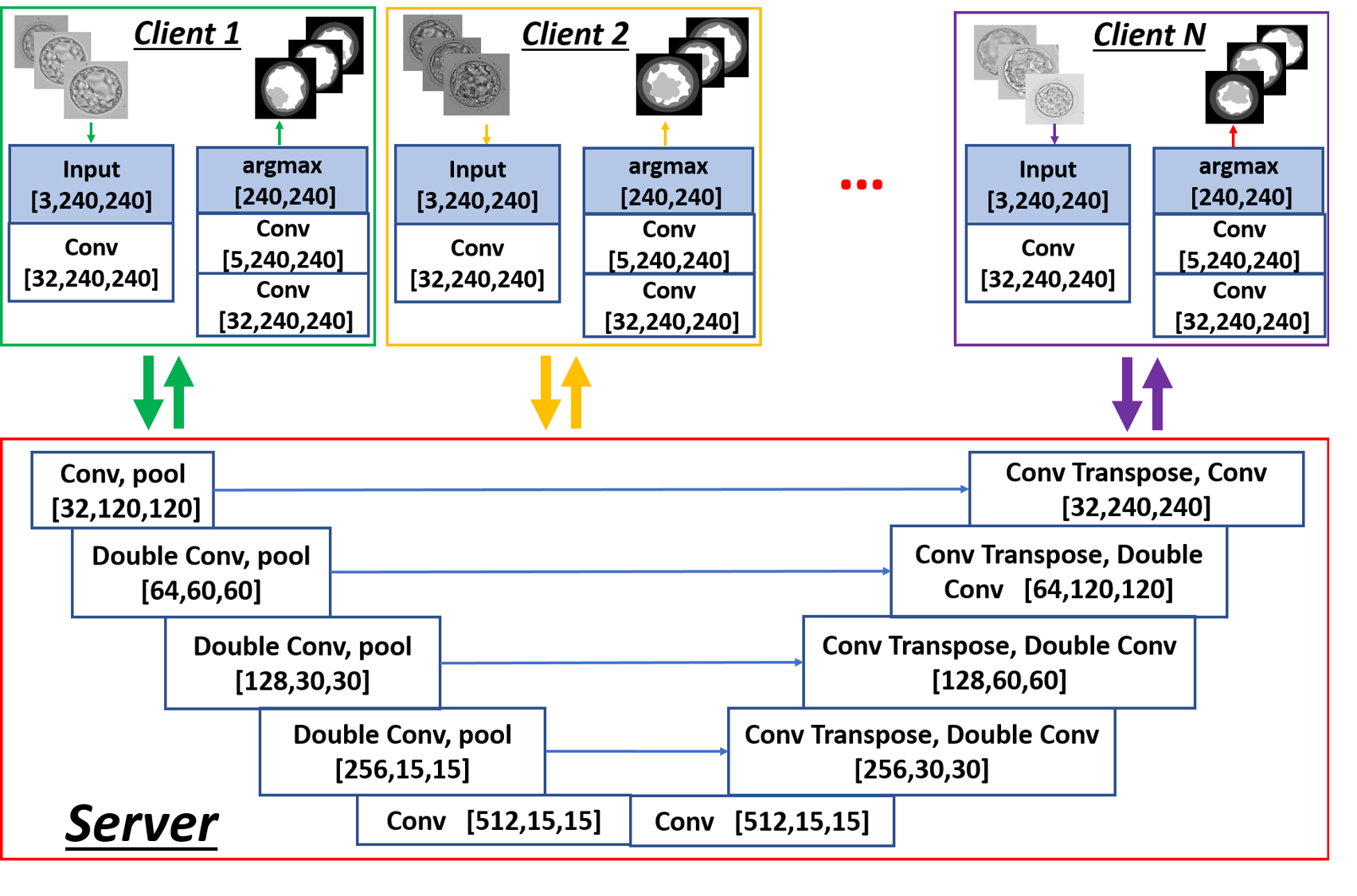}
    \vspace*{-3mm}
    \caption{\small{Our SplitFed architecture.}}
    \label{Fig:SplitFed_U-Net}
\end{figure}

We study training of a SplitFed U-Net, as in Fig.~\ref{Fig:SplitFed_U-Net}, for segmentation of human embryo components in microscopic images. The overall U-Net consists of five downsampling blocks between the input and the bottleneck, and five upsampling blocks between the bottleneck and the output. Each block contains two convolutional layers with the kernel size of $3\times3$, followed by batch normalization and \texttt{ReLU} activation. Each downsampling block ends with a $2\times2$ max-pooling layer; the number of filters in downsampling blocks is 32, 64, 128, 256, and 512, respectively. Each upsampling block starts with a $2\times2$ upsampling layer, followed by tanspose convolution layers; the number of filters in upsampling blocks is 512, 256, 128, 64, and 32, respectively. The last convolutional layer is followed by an output \texttt{argmax} layer. The model is split  
such that the first convolutional layer (front-end, FE) and the last two convolutional layers together with the output \texttt{argmax} layer (back-end, BE) are on the client side, while the rest of the model is on the server.

\subsection{Training procedure and averaging}
The system adopts sequential training, where the server trains with each of the $N$ clients in turn. At the end of this training process, both client and server store their weights at the server side. Let's $W^{\text{S}}_i$ be the server weights while training with client $i$, and $W^{\text{C}}_i=\{W^{\text{FE}}_i, W^{\text{BE}}_i\}$ be the weights of client $i$, which include both FE and BE weights, from the same epoch. One pass across all clients is called a \textit{global epoch}. After all clients get trained locally, $W^{\text{S}}_i$ and $W^{\text{C}}_i$ are averaged into $\overline{W}^\text{S}$ and $\overline{W}^\text{C}$, respectively, to create a global model of  $W=\{\overline{W}^\text{S}, \overline{W}^\text{C}\}$. $\overline{W}^\text{C}$ is sent back to all clients and local validation is performed on the global model. Considering the validation loss statistics, global model is updated one more time before starting the next global epoch. 

Within each local epoch, in the forward pass, features produced by FE are sent to the server. The server then processes them using its part of the model. The server features are sent to the client-side's BE. In the back-propagation pass, gradient updates from the BE are sent to the server, and back-propagated through the server-side part of the model to FE.  

Algorithm~\ref{alg:QA-SplitFed} describes the QA-SplitFed averaging method. Its inputs are: $N$ (the number of  clients), $G$ (the number of global epochs), $E$ (the number of local epochs), $\mathbf{d}^{t} = [m_1,...,m_N]^{\top}/(\sum_i{m_i})$ (the vector of training data scores), and $\mathbf{d}^{v} = [n_1,...,n_N]^{\top}/(\sum_i{n_i})$ (the vector of validation data scores). Its outputs are $\overline{W}^{\text{C}}$ (global client model) and $\overline{W}^{\text{S}}$ (global server model).

\begin{algorithm}[!t]
\caption{ QA-SplitFed  }
\label{alg:QA-SplitFed}

\DontPrintSemicolon

\textbf{Inputs:}\;
\quad $N$: Number of clients\;
\quad $G$: Number of global epochs\;
\quad $E$: Number of local epochs\;
\quad $\mathbf{d}^{t}$: Vector of training data scores\;
\quad $\mathbf{d}^{v}$: Vector of validation data scores\; 
\textbf{Outputs:}\;
\quad $\overline{W}^{\text{C}}= \{\overline{W}^{\text{FE}}, \overline{W}^{\text{BE}}\}$: Global client model \; 
\quad $\overline{W}^{\text{S}}$: Global server model \; 

\For{ $g=1$ \KwTo $G$}{

    \For{$i=1$ \KwTo $N$ } {
        Local training and validation over $E$ epochs \;
        $ W_i = \{ W^{\text{C}}_{i} , W^{\text{S}}_{i} \}  \leftarrow$ model's weights from epoch with the lowest validation loss\;   
        $ \mathcal{L}_i^j  \leftarrow$ loss of $W_i$ at training sample $j$  \; 
        $\mu_i^{t} \gets \texttt{mean}\{\mathcal{L}^1_i,...,\mathcal{L}^{m_i}_i\}$ \;
        $\sigma_i^{t} \gets \texttt{std}\{\mathcal{L}^1_i,...,\mathcal{L}^{m_i}_i\}$ \; 
        $b_i^{t} \gets \mu_i^{t} + 2\sigma_i^{t}$ \;
    }
    $ \mathbf{b}^{t} \gets [b_1^{t},...,b_N^{t}]^{\top} $ \;
    $ \mathbf{W} \gets [W_1,...,W_N]^{\top} $ \;
    $ \{\overline{W}^{\text{C}}, \overline{W}^{\text{S}} \} \leftarrow$ \texttt{ModelUpdates}($\mathbf{W}, \mathbf{b}^{t} , \mathbf{d}^{t} $) \; 
    
    \For{$i=1$ \KwTo $N$ } {
        $ \mathcal{L}_i^j  \leftarrow$ loss of $\{\overline{W}^{\text{C}}, \overline{W}^{\text{S}}\}$ at valid. sample $j$  \; 
        $\mu_i^{v} \gets \texttt{mean}\{\mathcal{L}^1_i,...,\mathcal{L}^{n_i}_i\}$ \;
        $\sigma_i^{v} \gets \texttt{std}\{\mathcal{L}^1_i,...,\mathcal{L}^{n_i}_i\}$ \; 
        $b_i^{v} \gets \mu_i^{v} + 2\sigma_i^{v}$ \;
    }
    $ \mathbf{b}^{v} \gets [b_1^{v},...,b_N^{v}]^{\top} $ \;
    $ \{\overline{W}^{\text{C}}, \overline{W}^{\text{S}} \} \leftarrow$ \texttt{ModelUpdates}($\mathbf{W}, \mathbf{b}^{v}, \mathbf{d}^{v} $) \; 
    } 
\textbf{function} \texttt{ModelUpdates} ($\mathbf{W}, \mathbf{b} , \mathbf{d}$)   \;
\quad  $\mathbf{q} \gets \texttt{softmax}[1 / \mathbf{b}] = [q_1,...,q_N]^{\top} $  \;
\quad  $\mathbf{r} \gets \frac{\mathbf{q}\odot\mathbf{d}}{\mathbf{q}^{\top}\mathbf{d}} = [r_1,...,r_N]^{\top}$  \;
\quad  $ \overline{W}^{\text{C}} \gets 
    \sum_{i=1}^N r_i W^{\text{C}}_i,\quad  \overline{W}^{\text{S}} \gets \sum_{i=1}^N r_iW^{\text{S}}_i$ \; 

\end{algorithm}

At each global epoch, after each client $i$  
is  trained locally, the mean training loss value $\mu_i^{t}$ at the best local epoch  
is calculated. To have a better understanding of the loss values and their distributions, the standard deviation $\sigma_i^{t}$ of the loss values at the best epoch are also calculated. The mean loss value based on a relatively few data samples is unreliable, so we take 
an upper bound of a $95\%$ confidence interval for the mean. If the loss values were normally distributed, this upper bout would be 
$b_i^t=\mu_i^{t}+2\sigma_i^{t}$. Hence, $b_i^t$ is an indicator of the (un)reliability of client $i$.

After one round of local training of all clients, two vectors, $\mathbf{b}^{t} = [b_1^{t},...,b_N^{t}]^{\top} $ and  $\mathbf{q}^t = \texttt{softmax}[1 / \mathbf{b}^t]$, are created. A smaller $b_i^t$,  
and thus a larger $q_i^t$, represents a more reliable client. In addition to $\mathbf{b}^t$ and $\mathbf{q}^t$, we use the training data score vector $\mathbf{d}^t = [m_1,...,m_N]^{\top} / \sum_{i}^{} m_i$, similar to~\cite{mcmahan2017communication, FedAVGM}, which specifies the training data distribution across clients.
By combining clients' training quality scores $\mathbf{q}^t$ and training data scores $\mathbf{d}^t$, averaging weights are calculated as $\mathbf{r}=\frac{\mathbf{q}^t\odot\mathbf{d}^t}{{\mathbf{q}^t}^{\top}\mathbf{d}^t}$, where $\odot$ is the Hadamard (element-wise) product. Note that $\sum_{i}r_i = 1$. Then, averaging is performed as:
$\overline{W}^{\text{S}} = \sum_{i=1}^N r_iW^{\text{S}}_i, \quad \overline{W}^{\text{C}} = \sum_{i=1}^N r_iW^{\text{C}}_i$.

$\overline{W}^{\text{C}}$ is broadcast to all clients  
and the server model is 
updated to $\overline{W}^{\text{S}}$. Local validation is performed using the new local and global models, to assess the performance on each client's validation data. Then the above averaging process is repeated, but with validation vectors $\mathbf{b}^v$ and $\mathbf{d}^v$ taking the place of $\mathbf{b}^t$ and $\mathbf{d}^t$.  
This process is repeated $G$ times. The best global model is chosen according to the global epoch with the smallest global validation loss.

\vspace{-10pt}
\section{Experiments}
\label{sec:experiments}
\vspace{-5pt}
\subsection{Dataset description and clients configuration setup}
\vspace{-5pt}
We train the SplitFed U-Net model (Fig.~\ref{Fig:SplitFed_U-Net}) on multi-label segmentation of human blastocyst components~\cite{blastnet,blastosys_data}.  
Our training dataset includes 815 embryo images, each with a segmentation mask image that includes four components: Zona Pellucida (ZP), Trophectoderm (TE), Inner Cell Mass (ICM), and Blastocoel (BL).
A test set of 100 images with uncorrupted GTs is saved for testing the global model. Our system includes 5 clients. Input images are distributed non-uniformly across clients as follows: client 1: 210, client 2: 120, client 3: 85, client 4:180 and client 5: 120, with each client using 85\% of it's images for training and 15\% for validation. 

\begin{figure}[t]
    \centering
    \includegraphics[width=0.7\columnwidth]{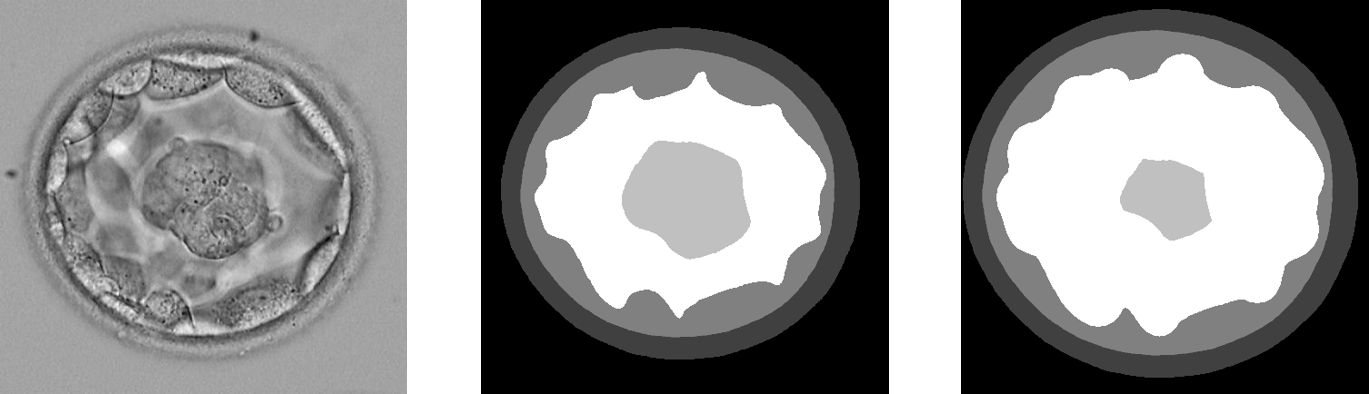}
    \caption{\small{Left to right: input image, GT mask, corrupted mask.}}
    \label{fig:mask_corruption}
\end{figure}

To corrupt a ground truth mask, mask segments were dilated by a circular structuring element of radius 20 pixels. The effect of this is that boundaries between  segments are shifted, emulating inaccurate segment boundary annotation.  
An example is given in Fig.~\ref{fig:mask_corruption}. 
For each corrupted client, corruption is applied to all its training and validation masks. 

\begin{table}[htbp]
\caption{\small{Comparison of baseline averaging methods with QA-SplitFed on human embryo test set. Acc. is the accuracy; the last four columns show \emph{Jaccard Indices} for ZP, TE, ICM, and BL.}}
\vspace{-0.5\baselineskip}
\label{tab:results}
\center
\setlength{\tabcolsep}{3 pt}
\renewcommand{\arraystretch}{1}
\footnotesize
\begin{tabular}{|c|c|c|c|c|c|c|c|}
\hline 

\textbf{ \shortstack{No. of \\ corrupted \\ clients} } &  \shortstack{\quad  \\ \textbf{Method}  \\ \quad \\ \quad \\ \quad  }  &  \shortstack{\quad  \\ \textbf{Loss}  \\ \quad \\ \quad \\ \quad   } &  \textbf{\shortstack{Acc. \\ $[\%] $} } &    \shortstack{\textbf{ZP} \\ $[\%] $}  &  \shortstack{\textbf{TE}  \\ $[\%] $}  & \shortstack{\textbf{ICM}  \\ $[\%] $}  & \shortstack{\textbf{BL} \\ $[\%] $}   \\ \hline \hline

\multirow{6}{*}  
 &  Naive &   0.10 & 91.87  &    74.21 &  73.54 &  82.35  &  86.19    \\ 
 &  FedAVG &   0.11  & 91.46  &    72.22  &  73.18  &  84.56 &   87.10  \\
&  FedAVG-M &   0.15  & 89.18  &   69.13  &  64.16  &  72.25 &   80.61  \\
 0  &  SplitAVG &   0.10  & 92.25  &    76.21  &  72.32  &  82.10 &   85.11  \\
 &  FedMix &   0.14  & 89.00 &   73.45  &  74.21  &  76.12 &   79.12  \\
 & QA-SplitFed  & \textbf{0.10} &  \textbf{93.28} &   \textbf{ 77.21} & \textbf{ 74.65} &  \textbf{ 83.47}  & \textbf{ 87.23}  \\
 \hline \hline 
\multirow{6}{*} 
 &  Naive &  0.13  &  90.31 &  68.04  &   68.62  &   80.43  &   85.22   \\ 
 &  FedAVG &  0.11  &  91.25 & 73.21  &   71.26  &   81.17  &   86.09 \\
 &  FedAVG-M &   0.18  & 87.67 & 65.22  &  62.55  &  62.69 &   78.37  \\
 1 &  SplitAVG &    49  & 66.37 &  16.25  &  13.76  &  58.07  &   67.01  \\
 &  FedMix &   0.15  & 88.34 &   68.03  &  71.11  &  75.18 &   81.15  \\
 & QA-SplitFed & \textbf{0.10} &  \textbf{92.52} &  \textbf{ 76.43} & \textbf{ 74.66}  &  \textbf{ 83.02}  & \textbf{ 87.17} \\
 \hline \hline

\multirow{6}{*}
 &  Naive &   0.26  &   79.79  &  40.22  &    68.35  &    73.11  &    77.13 \\ 
 &  FedAVG &  0.24  &  82.93  &    45.28  &   39.21  &   71.32  &   78.00 \\ 
 &  FedAVG-M &  0.25   & 83.24  &  59.65  &  37.87  &  57.18 &   71.09  \\
2  &  SplitAVG &    0.42  & 69.52&   18.21  &  17.76  &  52.02 &   68.18 \\
 &  FedMix &   0.23  & 79.24 &   51.50  &  63.21  &  69.36 &   74.12  \\
 & QA-SplitFed  &  \textbf{0.10} &   \textbf{91.98} &     \textbf{ 74.55}  &   \textbf{ 73.87} &   \textbf{ 83.65}  & \textbf{ 87.18} \\ 
\hline \hline

\multirow{6}{*}
 &  Naive &   0.34 &   75.19 &  29.80 &    21.25 &    63.35  &    71.29  \\
 &  FedAVG &   0.34  &  75.27  &   29.02  &   23.00  &   63.04  &   72.16   \\
 &  FedAVG-M &   0.33   & 77.45  &   47.26  &  19.65  &  58.33 &   67.72  \\
3  &  SplitAVG &   0.40  & 73.44  & 21.00  &  15.19  &  51.05 &   68.61  \\
 &  FedMix &   0.28  & 78.20  &   43.01  &  37.05  &  68.07 &   71.91  \\
 &  QA-SplitFed &  \textbf{0.10} &  \textbf{92.40} &  \textbf{ 75.05} &  \textbf{ 71.18} &  \textbf{ 81.27} & \textbf{ 86.98} \\
 \hline \hline

\multirow{6}{*}
 &  Naive &  0.45  &  65.50  &    16.76  &   11.35  &   53.87  &   67.46  \\
 &  FedAVG & 0.43  & 68.79  &   19.71  &  13.00  &  54.14  &  67.42 \\ 
 &  FedAVG-M &   0.56  & 63.95 & 28.63  &  14.08  &  0.00 &   58.08  \\
 4 &  SplitAVG &   0.59  & 62.74 &   17.09  &  13.23  &  0.00 &   60.87  \\
 &  FedMix &   0.50  & 65.44  &  17.85  &  12.84  &  49.02 &   63.00  \\
 & QA-SplitFed  &    \textbf{0.10} &  \textbf{92.00}   &     \textbf{ 75.08} &  \textbf{ 73.09} &  \textbf{ 81.25}  & \textbf{ 86.34 } \\ 
\hline \hline

\multirow{6}{*}
&  Naive &  0.50 &  61.50 &  12.22  &   7.08  &   42.21  &   62.45 \\ 
&  FedAVG &  0.50  &  60.75 &  12.01  &   8.12  &   45.22  &   63.42  \\
&  FedAVG-M &   0.64  & 57.78  &  15.16  &  3.08  &  0.00 &   54.39  \\
 5 &  SplitAVG &   \textbf{0.48}  & \textbf{62.35} &    \textbf{ 14.14}  & \textbf{ 11.81}  & \textbf{ 46.95} &  \textbf{ 64.10}  \\
 &  FedMix &   0.51  & 61.15  &   14.36  &  7.49  &  44.01 &   58.00 \\
&  QA-SplitFed  &    0.51  &  60.71 &     13.52 &   0.00 &   45.65 &  61.36 \\ 
\hline 

\end{tabular}
\vspace*{-\baselineskip}
\label{tab1}
\end{table} 

\begin{figure*}[!htb]
\center
\includegraphics[width=14cm, height = 5.5cm]{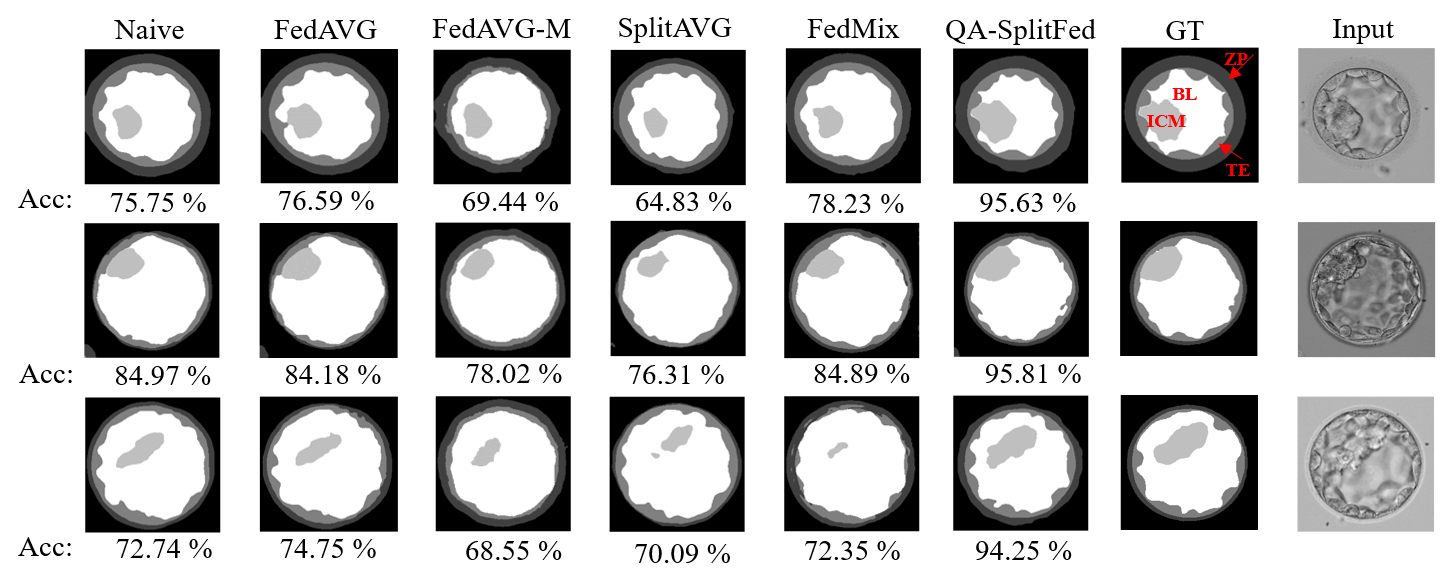}
    \caption{\small{Visual comparison of the predicted segmentation masks produced by various methods  with 
    3 out of 5 corrupted clients.}}
    \label{Fig:visual-results}
\end{figure*}

\subsection{Results and discussion}
\label{sec:results}
\vspace{-3pt}
We compare our QA-SplitFed against five baseline methods: Naive SplitFed ($\overline{W}=\frac{1}{N}\sum_{i=1}^N W_i$), FedAVG~\cite{mcmahan2017communication} ($\overline{W}=\frac{1}{\sum_{i}m_i}\sum_{i=1}^N m_iW_i$), FedAVG-M~\cite{FedAVGM} (SGD optimizer and momentum of $0.9$), SplitAVG~\cite{splitAVG}, and FedMix~\cite{fedmix} ($\beta=1.5 $, $\lambda=10$).  All models are trained and validated over $E=12$ local epochs and $G=10$ global epochs. 

\begin{figure}[tb]
    \centering
    \includegraphics[ width=8.6cm]{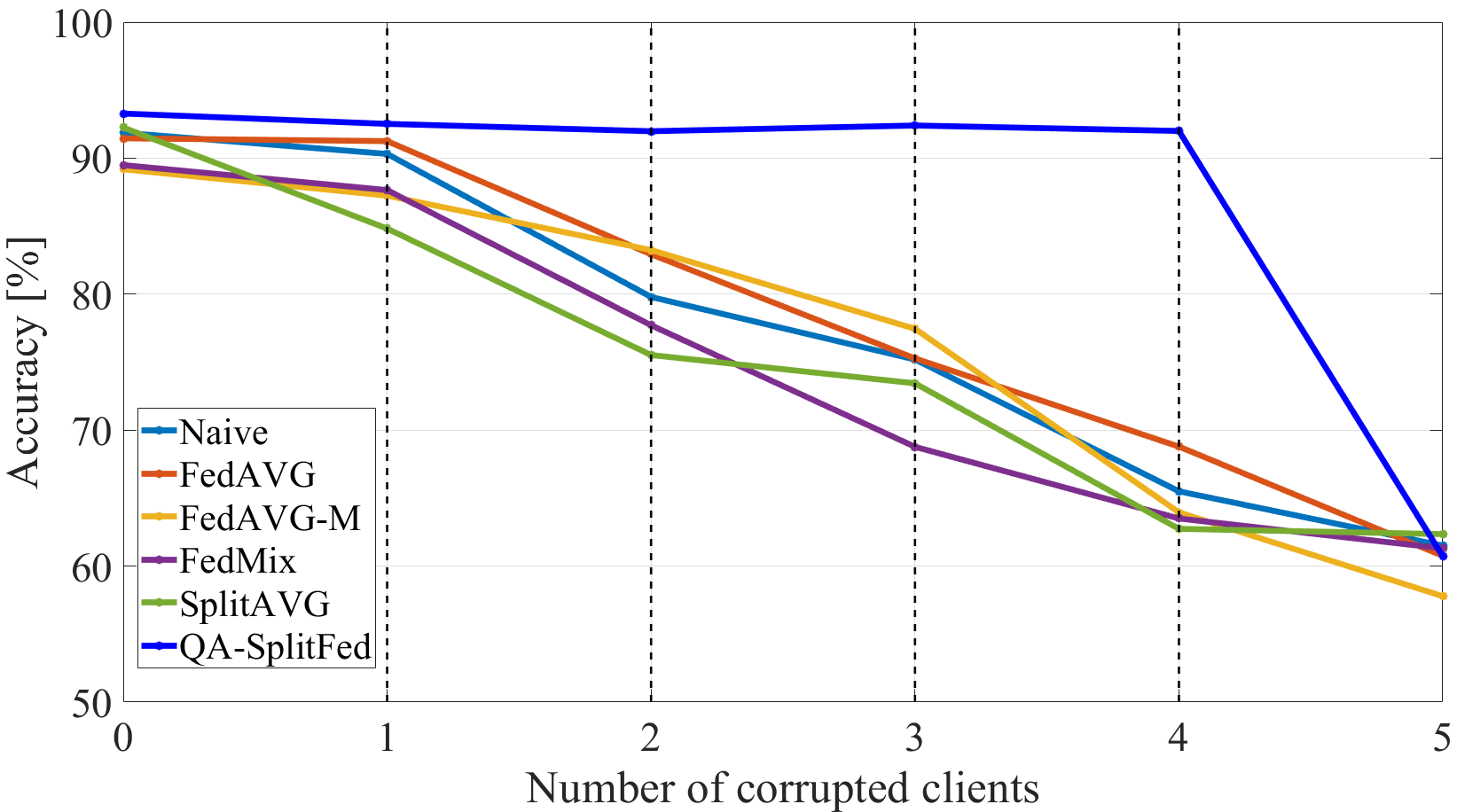}
    \vspace*{-3mm}
    \caption{\small{Global model Accuracy vs no. of corrupted clients.}}
    \label{Fig:comparison}
\end{figure}

The test set average results are shown in Table~\ref{tab:results}. In this table, accuracy (Acc.) is defined as the percentage of correctly classified pixels, and loss is the normalized global model loss. The reported values in ZP, TE, ICM and BL are \emph{Jaccard Indices}. Experiments were conducted for different numbers of corrupted clients as shown in first column of this table. Best results are in bold. These results show that the performance of all baseline methods drops dramatically as the number of corrupted clients increases. Our proposed method, QA-SplitFed, maintains its accuracy even with 4 out of 5 clients corrupted, 
so it is able to build an accurate global model with only one accurate client. When all clients are corrupted, all averaging methods lead to a sub-par global models. In this extreme case, SplitAVG had slightly better performance than other methods, but it still produced an inaccurate model. Fig.~\ref{Fig:comparison} shows the 
accuracy of the various methods 
for different numbers of corrupted clients. Fig.~\ref{Fig:visual-results} displays three test images with their corresponding correct GTs and their predicted masks 
with 3 out of 5 corrupted clients. As seen in the figure, the masks predicted by our method are closest to GTs.

\vspace{-5pt}
\section{Conclusion}
\label{sec:conclusions}
\vspace{-5pt}
In this paper we proposed a new model averaging method for split-federated learning in the presence of corrupted or poorly annotated clients. We showed that existing averaging methods  
are fairly fragile in that context. Our proposed method overcomes the challenge of corrupted clients by monitoring the  training and validation loss 
at each client and adjusting the averaging weights adaptively, leading to a more accurate global model even when all but one clients are corrupted. Moreover, QA-SplitFed is more scalable than FL since it can scale to less capable clients (i.e., clients only need to train/run FE and BE, not the full model). Meanwhile, aggregation complexity is similar to that of federated learning.

\bibliographystyle{IEEEbib}
\bibliography{refs}

\end{document}